\documentclass[final]{article}

\usepackage[numbers]{natbib}

\usepackage[final]{nips_2018}

\usepackage{booktabs}       
\usepackage{amsfonts}       
\usepackage{nicefrac}       
\usepackage{microtype}      
\usepackage{graphicx}       
\usepackage{bm}             
\usepackage{amsmath}        
\usepackage{algorithm}      
\usepackage{algorithmic}    

\graphicspath{ {./images/} }

\title{Continual Learning Augmented Investment Decisions}

\author{
  Daniel Philps 
  \\
  Rothko Investment Strategies\\ 
  Department of Computer Science\\
  City, University of London\\
  \texttt{daniel.philps@city.ac.uk} 
  \And
  Tillman Weyde \\
  Department of Computer Science\\
  City, University of London\\
  \texttt{t.e.weyde@city.ac.uk} 
   \And
  Artur d'Avila Garcez \\
  Department of Computer Science\\
  City, University of London\\
  \texttt{a.garcez@city.ac.uk} 
  \And
  Roy Batchelor \\
  Cass Business School \\
  City, University of London\\
  \texttt{r.a.batchelor@city.ac.uk} 
}

\begin{document}

\maketitle

\begin{abstract}
Investment decisions can benefit from incorporating an accumulated knowledge of the past to drive future decision making. We introduce Continual Learning Augmentation (CLA) which is based on an explicit memory structure and a feed forward neural network (FFNN) base model and used to drive long term financial investment decisions. 
We demonstrate that our approach improves accuracy in investment decision making while memory is addressed in an explainable way. Our approach introduces novel remember cues, consisting of empirically learned change points in the absolute error series of the FFNN. Memory recall is also novel, with contextual similarity assessed over time by sampling distances using dynamic time warping (DTW). 
We demonstrate the benefits of our approach by using it in an expected return forecasting task to drive investment decisions. In an investment simulation in a broad international equity universe between 2003-2017, our approach significantly outperforms FFNN base models. We also illustrate how CLA's memory addressing works in practice, using a worked example to demonstrate the explainability of our approach.

\textbf{Keywords}: Continual learning, explicit memory, neural network, investment, trading.
\end{abstract}

\section{Introduction}
Time-series are ubiquitous in modern human activity, not least in finance, 
and applying continual learning to financial time-series problems would be highly beneficial. However, existing approaches suffer from problems such as catastrophic forgetting, the opacity of implicit memory or outright complexity 
, arguably making them not well suited to the high magnitude decision making that tends to be required in finance. 
We introduce Continual Learning Augmentation (CLA), a time-series based memory approach that accumulates models of past states, which are recalled and balanced when states approximately reoccur.
We apply CLA to drive investment decisions in international equities markets, the results of which show improved investment performance compared to a feedforward neural network (FFNN) base model. (In further research we have found that CLA outperforms more traditional, linear models also). 
We also show a worked example of how CLA's memory addressing adds explainability.

The remainder of this paper is organized as follows. 
The next section reviews related work on memory modelling.
We then introduce CLA in Section three, describing how different states are identified and remembered, and how models of past states are recalled and applied, as well as how less relevant models are forgotten. 
In Section four, we report experiments with feedforward neural networks that show significant improvements with CLA.
In Section five, we illustrate how CLA enables interpretation in terms of references to past states.
Finally, we present our conclusions and directions for future work. 

\section{Related work}
Approaches developed for continual learning have been applied to many areas and have employed a wide range of techniques including gated neural networks \cite{Hochreiter_1997}\cite{Chung_Bengio_2014}, explicit memory structures \cite{Weston}, prototypical addressing\cite{Snell_2017}, weight adaptation \cite{Hinton_Distilling_2015}\cite{Sprechmann_2018} to name a few. Having addressed catastrophic forgetting \cite{French1999CatastrophicFI} more recent research has turned to solving second order problems, such as overheads of external memory structures \cite{Rae_2016_sparsereads}, problems with weight saturation \cite{Kirkpatrick_2017} and the complications of outright complexity \cite{DBLP:journals/corr/ZarembaS15}. While a number of memory approaches apply to sequential memory tasks \cite{Graves_14}\cite{graves2016hybrid} a far smaller number still have been focused specifically on time-series \cite{Kadous_TS_2002:}\cite{Graves:2006:CTC:1143844.1143891}\cite{Lipton_TS_Modeling}\cite{Thomas_TS_2017}. It is unclear how effective these approaches would be in dealing with long term continual learning of noisy, non-stationary time-series, commonly found in finance.

\subsection{Financial regimes as a memory concept}
Regime switching models and change point detection provide a simplified answer to identifying changing states in time-series with the major disadvantage that change points between regimes (or states) are notoriously difficult to identify out of sample  \cite{fabozzi2010quantitative} and existing econometric approaches are limited by long term, parametric assumptions in their attempts  \cite{Page_1957,Picard_1985,Sugiura_1994,Chib_1998,Engle_1999,Zhang_2010,Siegmund_2013}. 
There is also no guarantee that a change point in a time-series represents a significant change in the accuracy of an applied model, a more useful perspective for modelling different states. 
Another approach is to focus on the change in the absolute error of a model, aiming to capture as much information as possible regarding changes in the relation between independent and dependent variables  \cite{Yu_2007}. 
Different forms of residual change have been developed \cite{Brown_1975,Jandhyala_1986,Jandhyala_1989,MacNeilt_1985,Bai_1991}. 
However, most approaches assume a single or known number of change points in a series and are less applicable to a priori change points or multivariate series  \cite{Yu_2007}. 

\subsection{Memory models} 
Memory modelling approaches using external memory structures require an appropriate memory addressing mechanism (a way of storing and recalling a memory). 
Memory addressing is generally based on a similarity measure such as cosine similarity  \cite{Graves_14,Park_2017} kernel weighting  \cite{Vinyals_2016}, use of linear models  \cite{Snell_2017} or instance-based similarities, many using K-nearest neighbours \cite{Kaiser_2017} \cite{Sprechmann_2018}. However, these approaches are not obviously well suited to assessing the similarity of noisy and non-stationary, multivariate time-series. 
Euclidean distance offers a way to compare time-series but has a high sensitivity to the timing of data-points, something that has been addressed by dynamic time warping (DTW) \cite{Sakoe1978,Ding_2008}. 
However, DTW requires normalized data \cite{KeoghKasetty_2003a} and is also computationally expensive, although some mitigating measures have been developed (see \cite{Zhang:2017:DTW:3062405.3062585}). 

\section{Continual Learning Augmentation}

Continual learning augmentation (CLA) is a regression approach applied as a sliding window stepping forward through time, over input data of one or more time-series. 
The approach is initialized with a an empty memory structure, $\bm{M}$ and a FFNN base model, $\phi$, parameterized by $\theta_B$. 
The base model is applied to a multivariate input series, $\bm{X}$, with $K$ variables over $T$ time steps.  
The base model produces a forecast value $\hat{y_{t+1}}$ in each period as time steps forward. 
A \emph{remember} function, $j$, appends a new model-memory, $\bm{M_t^m}$, to $\bm{M}$, on a \emph{remember cue} defined by the change in the base model's absolute error at time point $t$. 
A \emph{recall} function $g$ balances a mixture of base model and model-memory forecasts.

\begin{figure}
\includegraphics[width=\textwidth]{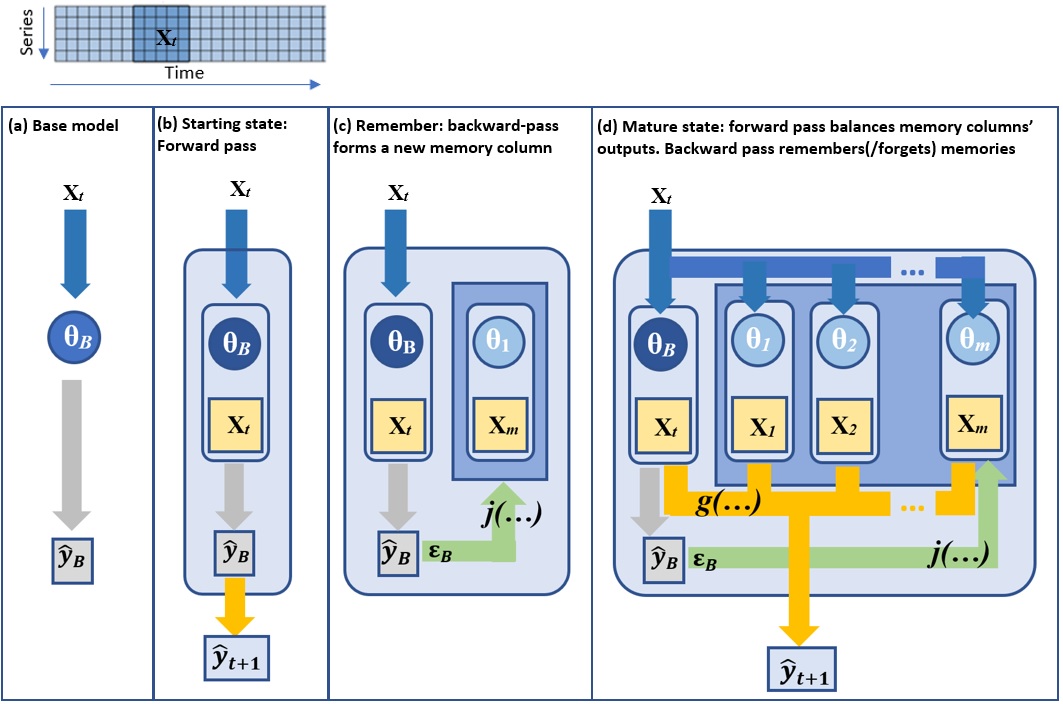}
\caption{Continual learning augmentation architecture. 
(a) A regression base model, $\phi$, parameterized by $\theta_B$, is run, stepping forward through time, training at each time step. (b) This base model is structured as a column containing model parameters, $\theta_B$, and a contextual reference, $\bm{X_t}$. 
Initially the base model is run as normal, completing a forward pass with the input data $\bm{X_t}$ to forecast ($\hat{y_{B,t+1}}$). 
(c) As time steps on, $y$ becomes observable and a backward pass is conducted where the absolute error of the base model, $\bm{\epsilon_B}$, determines if a change point has occurred,
On a change the function $j$ copies the base model column to a new memory column in $\bm{M}$.
The base model is then trained. 
(d) Over time more change points will be detected and more memory columns will be added. 
At each time step, all memory columns are run using the current input data $\bm{X_{t}}$. 
The forecast results of all the columns (including the base model column) are balanced by the function $g$ which uses the similarity of the current input, $X_{t-1}$ with the contextual reference of each memory, $\bm{X_{m}}$, to weight the output of each column, $\hat{y_{B}}$ and $\hat{y_{m}}$ of $\bm{M}$, to result in $\hat{y}_{(t+1)}$).}
\label{CLA_arch2}
\end{figure}
Figure \ref{CLA_arch2} shows the functional steps of remembering and recalling model-memories.

\subsection{Memory management} 
Repeating patterns are required in the input data to provide memory cues to remember and recall different past states. 
Model parameters trained in a given past state, $\theta_m$, can then be applied if that state approximately reoccurs in the future. 
When CLA forms a memory, it is stored as a column in an explicit memory structure, similar to \cite{Ciresan_2012}, which changes in size over time as new memories are remembered and old ones forgotten. Each memory column consists of a copy of a past base model parameterization, $\theta_{m}$, and the training data $\bm{X_{m}}$ used to learn those parameters; \begin{math}(\bm{X_{m}},\theta_{m})\end{math}.
As the sliding window steps into a new time period, CLA recalls one or more model-memories by comparing the latest input data ($\bm{X_{t}}$) with the training data stored in each memory column ($\bm{X_{m}}$). 
Memories with training data that are more similar to the current input series will have a higher weight applied to their output ($\hat{y}_{m,t+1}$) and therefore make a greater contribution to the final CLA output ($\hat{y}_{t+1}$).

\subsection{Remembering} 

Remembering is triggered by changes in the absolute error series of the base model, ${\bm{\epsilon_B}}$, as the approach steps forward through time. These changes are assumed to be associated with changes in state which are indicated by a function $j$. 
$j$ defines a change and stores a pairing of the parameterization of a base model, $\theta_B$, and a contextual reference, $\bm{X_{t}}$. 
Figure \ref{CLA_arch2}c) shows how a change is detected by function $j$ from a backward-pass, which then results in a new memory column being appended to $\bm{M}$: 
\begin{equation}\bm{M} = \big\{(\bm{X_{1}},\theta_1), \ldots ,(\bm{X_{m}},\theta_m)\big\}
\end{equation}

Immediately after the remember event has occurred, a new base model is trained on the current input, overwriting $\theta_B$. 

Theoretically, for a fair model of a state, $\bm{\epsilon_B}$ would be approximately $i.i.d.$ with a zero valued mean. Therefore the current base mode would cease to be a fair representation of the current state when $\bm{\epsilon_B}$ exceeds a certain confidence interval, in turn implying a change in state. $J_{Crit}$ represents a critical level for $\bm{\epsilon_B}$, indicating a change point has occurred in state. Memories are only stored when the observed absolute error series,$\bm{\epsilon_B}$, spikes above a critical level, $J_{Crit}$:


\begin{algorithm}
\caption{Remember function $j$}
\begin{algorithmic} 
\REQUIRE Initialize memory structure $\bm{M}$ 
\REQUIRE Initialize $J_{Crit}$
\STATE \# Step through time, period by period, starting at the earliest date 
\FOR{all time steps $t$ in $T$}
\STATE $\epsilon_{B,t}$ $\leftarrow$ error after forward pass of base model $\phi(\bm{X_t},\theta_{B})$
\IF{\begin{math}|\epsilon_{B,t}| \geq J_{Crit}\end{math}}
\STATE append model-memory $(\bm{X_t},\theta_{B})$ to $\bm{M}$
\ENDIF
\STATE \# Dependent variable becomes observable
\STATE $\theta_{B}$ $\leftarrow$ train base model and overwrite  $(\bm{X_t},\theta_{B})$ 
\STATE $J_{Crit}$ $\leftarrow$ learn and update $J_{Crit}$ 
\ENDFOR
\end{algorithmic}
\end{algorithm}

$J_{Crit}$ is a hyperparameter, optimized at every time step, to result in a level of sensitivity to remembering that forms an external memory, $\bm{M}$ , resulting in the lowest empirical forecasting error for the CLA approach over the study term up until time $T$:

\begin{center}
\begin{equation}
J_{Crit} =  
\underset{j_{Crit} \in jgrid} {\operatorname{argmin}}
f(\bm{X_t}, j_{Crit})
\end{equation}
\end{center}
Where $f$ is the CLA approach expressed as a function of the input series and $j_{Crit}$, yielding $\epsilon_{B,t}$ (the absolute error of the base model at time $t$). $jgrid$ is a 20 point, equidistant set between the minimum and the maximum values of $\bm{\epsilon_{B}}$.

\subsection{Recall} 
The recall of memories takes place in 
the function $g$,  
which calculates $\hat{y}_{m,(t+1)}$ 
a mixture of the 
predictions %
from the current base model and from model-memories.  
\begin{center}
\begin{equation}
\hat{y}_{(t+1)} = 
g(\bm{X_{t}},\bm{M^t})
\end{equation}
\end{center}
The mixture coefficients are 
based on comparing the similarity of the current time varying context $\bm{X_{t}}$ with the contextual references $\bm{X_{m}}$ stored with each individual memory. 
Memories that are more similar to the current context have a greater weight in CLA's final modelling outcome. 
%
Dynamic time warping (DTW) is used to calculate contextual similarity. 
However, multivariate DTW is computationally expensive \cite{Seto_2015}. 
As well as applying traditional constraints to the warping path, we also use a sampling based implementation to reduce expense further. 
DTW is only applied to a subset of $N$ randomly sampled instances from $\bm{X_{m}}$ and $\bm{X_{t}}$, sampling over rows, each of which represent different securities in the dataset:

\begin{center}
\begin{equation} \hat{D}(\bm{X_m},\bm{X_t}) = 1/N \sum_{i=0}^{N}DTW(X_{m,r_1(D)},X_{t,r_2(D)}) 
\end{equation}
\end{center}

Where $\hat{D}$ is the expected distance, $N$is the number of samples to take and $r_1(D), r_2(D)$ are random integers between 1 and $D$.

The mean, sampled distance is used to determine the similarity between the current context and those of each memory.

\subsection{Balancing} 
Two different approaches to memory balancing were used, firstly, the \emph{best individual} (i.e. lowest distance) model-memory:
\begin{equation}\hat{y}_{t+1}=g_{Best}(\bm{X_{t}}, \bm{M_{t}^{m}})
\end{equation}
where $g_{Best}$ is an output function to select the model-memory which is most similar to the current context (i.e. $\underset{m}{\operatorname{argmin }} \hat{D}(\bm{X_m},\bm{X_t})$),
 $\hat{y}_{t+1}$ is the regression output. Secondly, a \emph{similarity weighted} ensemble of all model-memories, $g_{SimWeight}(\bm{X_{t}}, \bm{M})$:  
\begin{equation}\hat{y}_{t+1}=\sum_{m=1}^{M}\phi(\bm{X_{t}},\theta_{m})\cdot\bigg[1-
\frac{\hat{D}(\bm{X_{m}},\bm{X_{t}})}{\sum_{m=1}^{M}\hat{D}(\bm{X_{m}},\bm{X_{t}})}
\bigg]
\end{equation}
Where $M$ is the number of memories in the memory structure $\bm{M}$. As a past state is unlikely to perfectly repeat, a continuous function for balancing model-memories is more likely to generalize better \cite{graves2016hybrid} than picking the best single model (which is indeed found to be the case).

\section{Simulating investment decisions 
}

\begin{figure}[t]
\includegraphics[width=\textwidth]{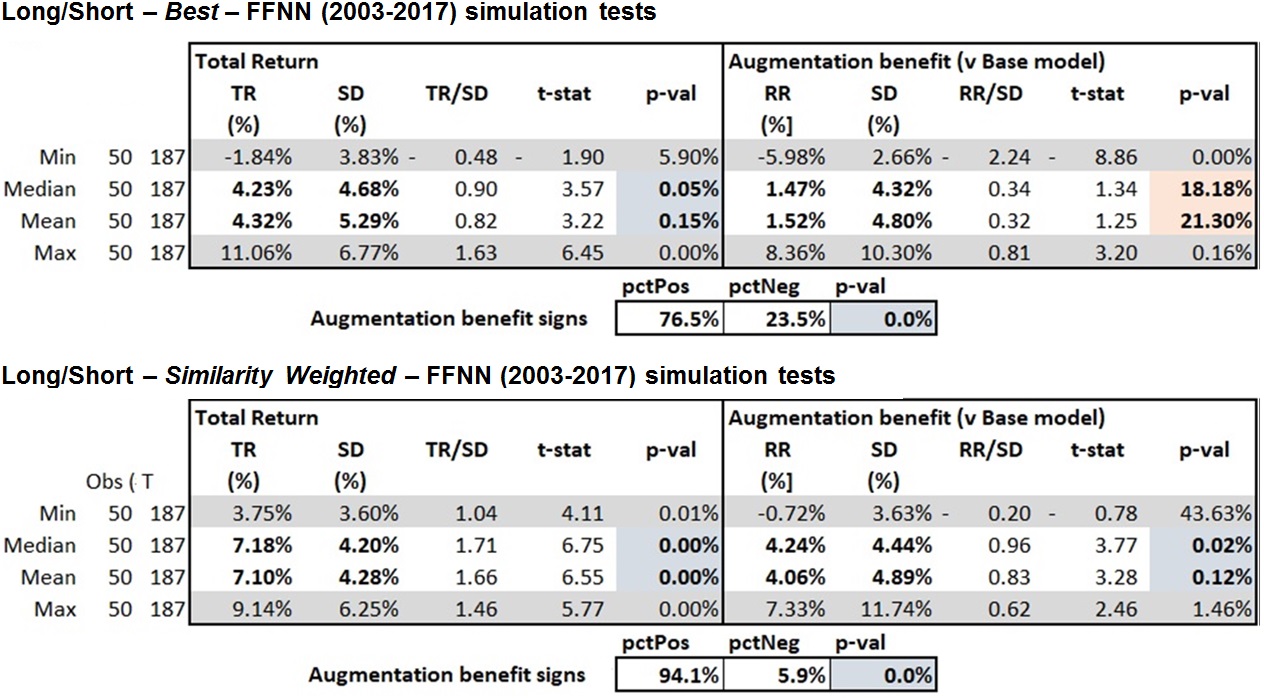}
\caption{Long/short investment simulations.
50 simulations were run using a CLA with $g_{best}$ (\emph{best}) balancing and then using $g_{SimWeight}$ (\emph{similarity weighted}). 
For each test the min, max, mean and median total return results are shown. 
TR is the annualized total return, SD the annualized standard deviation, TR/SD is the Sharpe ratio. 
Augmentation benefit shows the performance of CLA over the base model where RR is the annualized relative return of CLA over the base model and RR/SD is the information ratio. 
P-values of the t-stats of the Sharpe and information ratios show that CLA produces statistically significant Sharpe at the 1\% level, for the mean and median results 
and statistically significant information ratios, at the 1\% level, for the \emph{similarity weighted} simulations. 
The signs tests show statistically significant hit rates at the 1\% level for both \emph{best} and \emph{similarity weighted} tests.}
\label{LS_Sim}
\end{figure}

CLA is used in a regression task to forecast future expected returns of individual equity securities and used to drive an equities investment simulation. 
Stock level characteristics were used as the input dataset, to batch train an FFNN over all stocks in each period, forecasting US\$ total returns 12months ahead for each stock. Where a forecast was in the top(bottom) decile it was interpreted as a buy(sell) signal. 
\par
Although CLA is designed to use non-traditional driver variables, stock level characteristics are commonly expressed using factor loadings \cite{Fama93commonrisk}. To provide a more traditional context for testing, factor loadings were used as input data. These were estimated, in-sample at each time step by regressing style factor excess returns against each stock level US\$ excess return stream: 
$r_{i,t} = \alpha_{i,t} + \beta_{MKT,i,t}x_{MKT,t} + \beta_{VAL,i,t} x_{VAL,i,t} + \epsilon_{i,t}$,
where $r_{i,t}$ is the excess return of stock $i$ in period $t$, $x_{MKT,t}$ is the excess return of the All Countries World Ex-USA Equities Index, $x_{VAL,t}$ is the relative return of the All Countries World Ex-USA Value Equities Index. 
\par
Stock level factor loadings populated a matrix, $\bm{X}$, which comprized the input data. 
Each row represented a stock appearing in the index at time $t$ (up to 4,500 stocks) and each column related to a coefficient calculated on a specific time lag. 
$\bm{X}$ resulted from winsorizing the raw input to eliminate outliers. 
A FFNN was trained in each period by separating the input data into training, cross validation and testing sets in 75/5/25 proportions.
Long/short model portfolios were constructed every six months over the study term, simulating a rebalance every 6months, using equal weighted long positions (buys) and shorts (sells). 
The simulation encompassed 4,500 international equities in total, covering over 30 countries across developed and emerging markets, corresponding to the the All Countries World Ex-USA Equities Index between 2001-2017.
(Note that the first 24months were used as a training period while testing, which was entirely out of sample and free from known data snooping biases, started in 2003).
To account for the DTW sampling approach used, multiple test runs were carried out. 50 simulations were run per test for this purpose. Both \emph{best} and separately \emph{similarity weighted} were tested. 
Further testing was conducted to investigate whether results exhibited only an ensemble effect. 
An equal weighted balancing approach was also tested and found to generate weaker positive total returns relative to both \emph{best} and the \emph{similarity weighted} balancing approaches, demonstrating that CLA is exhibiting more than an ensemble effect

\section{Simulation results} 
\subsection{Accuracy: Investment decisions}

CLA results for long/short simulations showed a significant return benefit over the FFNN base model. 
(in further testing it was also found CLA produces positive and statistically significant augmentation benefits over traditional linear models also.)
It is also found that CLA is particularly effective at identifying stocks that produce a poor future return. 
As well as producing good performance relative to the base model, CLA also produced strong positive hit rates, statistically significant to the 5\% level or better in both tests.     
Examining the distribution of simulation results it is notable that as the base model error increases, augmentation benefit also increases, indicating a stronger augmentation benefit when the base model error is higher and vice versa. 
This property was exhibited in all tests conducted.

\subsection{Explainability: Interpretable Memory}
\begin{figure}
\begin{center}
\includegraphics[width=0.9\textwidth]{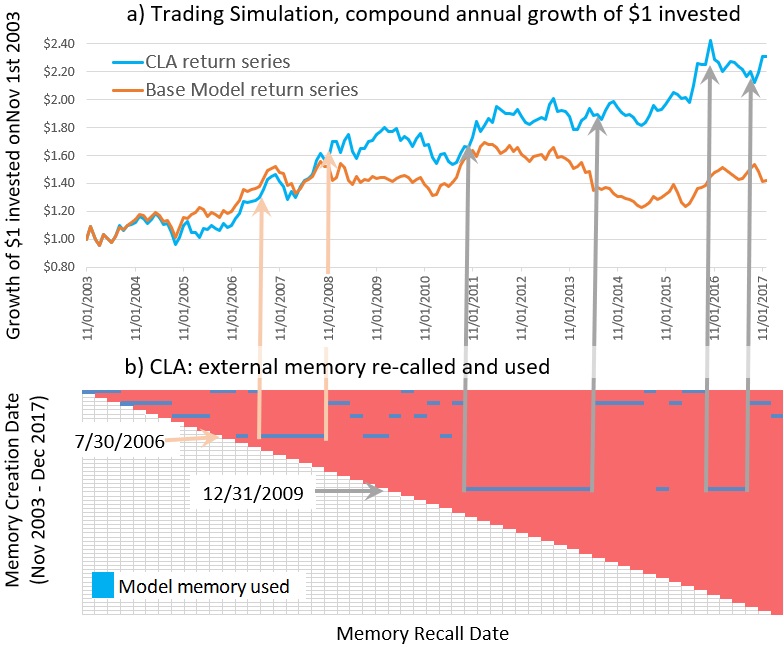}
\caption{Interpretable memory: how recalled memories contribute to simulation performance. 
The graph in a) uses
a single simulation and shows the growth of a \$1 investment in 2003, using strategies driven by CLA or the base model. 
b) shows a representation of the CLA memory structure, where each row in the expanding triangle represents a potential memory. 
This external memory structure can grow by one memory at each step forward in the simulation, although in practice only four memories were remembered in this simulation. 
The top row in the triangular graphic represents the base model. 
The memory with the highest weight in each period is highlighted.}
\label{CLAmem}
\end{center}
\end{figure}

CLA produces results that can be explained by examining which past models have been applied to which approximately repeating states, and with further investigation, why. 
Figure \ref{CLAmem} shows an example of a simulation run, where \ref{CLAmem}a) shows how the value of \$1 would have changed if invested in an investment strategy driven by CLA and, separately, by the base model. 
Figure \ref{CLAmem}b) shows the memory structure of CLA, where a new memory can theoretically be appended at every step forwards in the simulation, although only four memories were remembered in this example. 
Two memories are examined. Firstly, a memory formed in the period ending July 2006, which is used by CLA to outperform the base model in the period Sep 2007 - Oct 2008. 
Interestingly this is over the period of the \emph{Quant Quake} and up until just after the collapse of Lehman Brothers in the thick of the 2008 Financial crisis.
Secondly, a memory formed in the period ending December 2009 is used by CLA between 2011 and mid 2014, the period affected by the euro zone crisis and subsequent recovery. 
It is also used by CLA, albeit to far lesser effect, in late 2016.

\section{Conclusion}
Continual Learning Augmentation (CLA) is able to accumulate knowledge of changing market states over time and is able to apply this knowledge in approximately reoccurring future states. 
Our aim has been to combine different machine learning concepts to create a memory structure that improves the accuracy of time-series based modelling and produces memory modelling outcomes that are explainable. 
CLA is to our knowledge the first approach successfully applied to continual learning in the noisy, non-stationary, finance problem space, using an explicit memory approach to drive state dependent decision making. 
We introduce absolute error change as a memory concept, using the hyperparameter $J_{Crit}$ which learns points that govern remembering and forgetting of model-memories. 
We also use a sampling approach on multivariate DTW as a similarity measure to access CLA’s memory structure and make the first application of a memory modelling approach to a broad stock selection problem.

\subsection{Accuracy of outcomes}
We find that CLA produces positive, statistically significant forecasting benefit using a FFNN base model. Long/short tests show positive and statistically significant total returns and a positive and statistically significant augmentation benefit relative to the base model.
The similarity weighting of model-memories produces stronger results than simply picking the \emph{best} model-memory. 
If CLA were exploited in practice, this outperformance would give significant advantage to investment strategy returns. 

\subsection{Explainability of memory use}
CLA's memory structure can be interpreted in terms of which past state is relevant to forecasting in the current state.
This allows objective comparisons to be made between relevant past states and the current state and also allows for a better understanding of the characteristics of the current state in the context of similar past states. 
This information can provide deep insights to users to guide decision making. 

\subsection{Future work}
Our results indicate that CLA may be effectively applied to other problems on noisy and non-stationary time-series, in and outside of the finance domain. 
While our approach is directly applicable to quantitative investment we intend for our research to also be applied to other fields such as computational biology and wearable technology. 
It is also noted than the nature of absolute error change as a memory concept, introduced in this study, could hold more benefits for memory augmentation or model selection, where change in the absolute error \textit{distribution} could be used to better identify changing states and to better learn more appropriate models. 
We feel the benefits of our approach that we established in terms of accuracy and explainability, justify the additional complexity of CLA and warrant research into a fully differentiable structure for learning the relationships between memory cues, memory models and ultimate actions. 



\begin{thebibliography}{10}

\bibitem{Brown_1975}
Evans~J Brown~R, Durbin~J.
\newblock Techniques for testing the constancy of regression relationships over
  time.
\newblock {\em Journal of the Royal Statistical Society Series B
  (methodological)}, 37(2):149--192, 1975.

\bibitem{Chung_Bengio_2014}
Junyoung Chung, {\c{C}}aglar G{\"{u}}l{\c{c}}ehre, KyungHyun Cho, and Yoshua
  Bengio.
\newblock Empirical evaluation of gated recurrent neural networks on sequence
  modeling.
\newblock {\em CoRR}, abs/1412.3555, 2014.

\bibitem{Ciresan_2012}
Dan~C. Ciresan, Ueli Meier, and Jurgen Schmidhuber.
\newblock Multi-column deep neural networks for image classification.
\newblock {\em CoRR}, abs/1202.2745, 2012.

\bibitem{Picard_1985}
Picard D.
\newblock Testing and estimating change-points in time series.
\newblock {\em Advances in applied probability}, 1985.

\bibitem{Ding_2008}
Hui Ding, Goce Trajcevski, Peter Scheuermann, Xiaoyue Wang, and Eamonn Keogh.
\newblock Querying and mining of time series data: Experimental comparison of
  representations and distance measures.
\newblock In {\em Proceedings of the VLDB Endowment}, volume~1, pages
  1542--1552. Wiley, 8 2008.

\bibitem{Page_1957}
Page E.
\newblock On problems in which a change in a parameter occurs at an unknown
  point.
\newblock {\em Biometrika}, 44(1/2):248--252, 1957.

\bibitem{Engle_1999}
Smith~A Engle~R.
\newblock Stochastic permanent breaks.
\newblock {\em The Review of Economics and Statistics}, 81(4):553--574, 1999.

\bibitem{fabozzi2010quantitative}
F.J. Fabozzi, S.M. Focardi, and P.N. Kolm.
\newblock {\em Quantitative Equity Investing: Techniques and Strategies}.
\newblock Frank J. Fabozzi Series. Wiley, 2010.

\bibitem{Fama93commonrisk}
Eugene~F. Fama and Kenneth~R. French.
\newblock Common risk factors in the returns on stocks and bonds.
\newblock {\em Journal of Financial Economics}, 33:3--56, 1993.

\bibitem{Thomas_TS_2017}
Thomas Fischer and Christopher Krauss.
\newblock Deep learning with long short-term memory networks for financial
  market predictions.
\newblock FAU Discussion Papers in Economics 11/2017, Friedrich-Alexander
  University Erlangen-Nuremberg, Institute for Economics, 2017.

\bibitem{French1999CatastrophicFI}
French.
\newblock Catastrophic forgetting in connectionist networks.
\newblock {\em Trends in cognitive sciences}, 3 4:128--135, 1999.

\bibitem{Graves:2006:CTC:1143844.1143891}
Alex Graves, Santiago Fern\'{a}ndez, Faustino Gomez, and J\"{u}rgen
  Schmidhuber.
\newblock Connectionist temporal classification: Labelling unsegmented sequence
  data with recurrent neural networks.
\newblock In {\em Proceedings of the 23rd International Conference on Machine
  Learning}, ICML '06, pages 369--376, New York, NY, USA, 2006. ACM.

\bibitem{Graves_14}
Alex Graves, Greg Wayne, and Ivo Danihelka.
\newblock Neural turing machines.
\newblock {\em CoRR}, abs/1410.5401, 2014.

\bibitem{graves2016hybrid}
Alex Graves, Greg Wayne, Malcolm Reynolds, Tim Harley, Ivo Danihelka, Agnieszka
  Grabska-Barwińska, Sergio~Gómez Colmenarejo, Edward Grefenstette, Tiago
  Ramalho, John Agapiou, Adrià~Puigdomènech Badia, Karl~Moritz Hermann, Yori
  Zwols, Georg Ostrovski, Adam Cain, Helen King, Christopher Summerfield, Phil
  Blunsom, Koray Kavukcuoglu, and Demis Hassabis.
\newblock Hybrid computing using a neural network with dynamic external memory.
\newblock {\em Nature}, 538(7626):471--476, October 2016.

\bibitem{Yu_2007}
Yu~H.
\newblock High moment partial sum processes of residuals in arma models and
  their applications.
\newblock {\em Journal of time series analysis}, 28(1):72--91, 2007.

\bibitem{Hinton_Distilling_2015}
Geoffrey Hinton, Oriol Vinyals, and Jeffrey Dean.
\newblock Distilling the knowledge in a neural network.
\newblock In {\em NIPS Deep Learning and Representation Learning Workshop},
  2015.

\bibitem{Hochreiter_1997}
Schmidhuber~J Hochreiter~S.
\newblock Long short term memory.
\newblock {\em Neural Computation}, 9(8):1735--1780, 1997.

\bibitem{Bai_1991}
Bai J.
\newblock On the partial sums of residuals in autoregressive and moving average
  models.
\newblock {\em Journal of Timeseries Analysis}, 14(3), 1991.

\bibitem{Jandhyala_1989}
V.~K. Jandhyala and I.~B. MacNeill.
\newblock Residual partial sum limit process for regression models with
  applications to detecting parameter changes at unknown times.
\newblock {\em Stochastic Processes and their Applications}, 33(2):309--323,
  1989.

\bibitem{Jandhyala_1986}
MacNeill~I Jandhyala~V.
\newblock The change point problem: a review of applications.
\newblock {\em Developments in water science}, 27:381--387, 1986.

\bibitem{Kadous_TS_2002:}
M.W. Kadous, Mohammed~Waleed Kadous, and Supervisor~Claude Sammut.
\newblock Temporal classification: Extending the classification paradigm to
  multivariate time series.
\newblock Technical report, Unknown, 2002.

\bibitem{Kaiser_2017}
Lukasz Kaiser, Ofir Nachum, Aurko Roy, and Samy Bengio.
\newblock Learning to remember rare events.
\newblock {\em CoRR}, abs/1703.03129, 2017.

\bibitem{KeoghKasetty_2003a}
Eamonn~J. Keogh and Shruti Kasetty.
\newblock On the need for time series data mining benchmarks: {A} survey and
  empirical demonstration.
\newblock {\em Data Min. Knowl. Discov.}, 7(4):349--371, 2003.

\bibitem{Kirkpatrick_2017}
James Kirkpatrick, Razvan Pascanu, Neil~C. Rabinowitz, Joel Veness, Guillaume
  Desjardins, Andrei~A. Rusu, Kieran Milan, John Quan, Tiago Ramalho, Agnieszka
  Grabska{-}Barwinska, Demis Hassabis, Claudia Clopath, Dharshan Kumaran, and
  Raia Hadsell.
\newblock Overcoming catastrophic forgetting in neural networks.
\newblock {\em CoRR}, abs/1612.00796, 2016.

\bibitem{Lipton_TS_Modeling}
Zachary~Chase Lipton, David~C. Kale, Charles Elkan, and Randall~C. Wetzel.
\newblock Learning to diagnose with {LSTM} recurrent neural networks.
\newblock {\em CoRR}, abs/1511.03677, 2015.

\bibitem{MacNeilt_1985}
Ian~B. MacNeilt.
\newblock Detecting unknown interventions with application to forecasting
  hydrological data.
\newblock {\em JAWRA Journal of the American Water Resources Association},
  21(5):785--796, 1985.

\bibitem{Park_2017}
Seongsik Park, Sei~Joon Kim, Seil Lee, Ho~Bae, and Sungroh Yoon.
\newblock Quantized memory-augmented neural networks.
\newblock {\em CoRR}, abs/1711.03712, 2017.

\bibitem{Rae_2016_sparsereads}
Jack~W. Rae, Jonathan~J. Hunt, Tim Harley, Ivo Danihelka, Andrew~W. Senior,
  Greg Wayne, Alex Graves, and Timothy~P. Lillicrap.
\newblock Scaling memory-augmented neural networks with sparse reads and
  writes.
\newblock {\em CoRR}, abs/1610.09027, 2016.

\bibitem{Chib_1998}
Chib S.
\newblock Estimation and comparison of multiple change point models.
\newblock {\em Journal of Econometrics}, 86(2):221--241, 1998.

\bibitem{Sakoe1978}
H~Sakoe and S~Chiba.
\newblock Dynamic-programming algorithm optimization for spoken word
  recognition.
\newblock {\em IEEE TRANSACTIONS ON ACOUSTICS SPEECH AND SIGNAL PROCESSING},
  26(1):43--49, 1978.

\bibitem{Seto_2015}
Skyler Seto, Wenyu Zhang, and Yichen Zhou.
\newblock Multivariate time series classification using dynamic time warping
  template selection for human activity recognition.
\newblock {\em CoRR}, abs/1512.06747, 2015.

\bibitem{Siegmund_2013}
D~Siegmund.
\newblock Change-points: from sequential detection to biology and back.
\newblock {\em Sequential Analysis}, 32(2-14):43--46, 2013.

\bibitem{Snell_2017}
Jake Snell, Kevin Swersky, and Richard~S. Zemel.
\newblock Prototypical networks for few-shot learning.
\newblock {\em CoRR}, abs/1703.05175, 2017.

\bibitem{Sprechmann_2018}
Pablo Sprechmann, Siddhant~M. Jayakumar, Jack~W. Rae, Alexander Pritzel,
  Adri{\`{a}}~Puigdom{\`{e}}nech Badia, Benigno Uria, Oriol Vinyals, Demis
  Hassabis, Razvan Pascanu, and Charles Blundell.
\newblock Memory-based parameter adaptation.
\newblock {\em CoRR}, abs/1802.10542, 2018.

\bibitem{Sugiura_1994}
Ogden~R Sugiura, N.
\newblock Testing change-points with linear trend.
\newblock {\em Communications in Statistics B: Simulation and Computation},
  23:287--322, 1994.

\bibitem{Vinyals_2016}
Oriol Vinyals, Charles Blundell, Timothy~P. Lillicrap, Koray Kavukcuoglu, and
  Daan Wierstra.
\newblock Matching networks for one shot learning.
\newblock {\em CoRR}, abs/1606.04080, 2016.

\bibitem{Weston}
Jason Weston, Sumit Chopra, and Antoine Bordes.
\newblock Memory networks.
\newblock {\em CoRR}, abs/1410.3916, 2014.

\bibitem{DBLP:journals/corr/ZarembaS15}
Wojciech Zaremba and Ilya Sutskever.
\newblock Reinforcement learning neural turing machines.
\newblock {\em CoRR}, abs/1505.00521, 2015.

\bibitem{Zhang:2017:DTW:3062405.3062585}
Zheng Zhang, Romain Tavenard, Adeline Bailly, Xiaotong Tang, Ping Tang, and
  Thomas Corpetti.
\newblock Dynamic time warping under limited warping path length.
\newblock {\em Inf. Sci.}, 393(C):91--107, July 2017.

\bibitem{Zhang_2010}
Ji~Hanlee Zhang~N, Siegmund~D and Li~J.
\newblock Detecting simultaneous change-points in multiple sequences.
\newblock {\em Biometrika}, 97:631--646, 2010.

\end{thebibliography}

\end{document}